\documentclass[conference]{IEEEtran}
\usepackage{times}

\usepackage[numbers]{natbib}
\usepackage{multicol}
\usepackage[bookmarks=true]{hyperref}
\usepackage{lipsum}
\usepackage{multirow}
\usepackage[table]{xcolor}
\usepackage{amsmath}

\usepackage{graphicx}
\usepackage{caption}
\usepackage{authblk}
\usepackage{hyperref}

\begin{document}

\title{Grounding Language Models with \\Semantic Digital Twins for Robotic Planning}

\author{
    Mehreen Naeem\textsuperscript{1},
    Andrew Melnik\textsuperscript{1},
    Michael Beetz\textsuperscript{1}.
}
\affil{\textsuperscript{1}Institute for
Artificial Intelligence, University of Bremen, Germany}
\affil{Corresponding author: \texttt{mnaeem@uni-bremen.de}}

\date{}  %

\maketitle

\begin{abstract}

We introduce a novel framework that integrates Semantic Digital Twins (SDTs) with Large Language Models (LLMs) to enable adaptive and goal-driven robotic task execution in dynamic environments. The system decomposes natural language instructions into structured action triplets, which are grounded in contextual environmental data provided by the SDT. This semantic grounding allows the robot to interpret object affordances and interaction rules, enabling action planning and real-time adaptability. In case of execution failures, the LLM utilizes error feedback and SDT insights to generate recovery strategies and iteratively revise the action plan. We evaluate our approach using tasks from the ALFRED benchmark, demonstrating robust performance across various household scenarios. The proposed framework effectively combines high-level reasoning with semantic environment understanding, achieving reliable task completion in the face of uncertainty and failure.

\end{abstract}

\begin{figure*}[thpb]
      \centering
      \includegraphics[width=\textwidth]{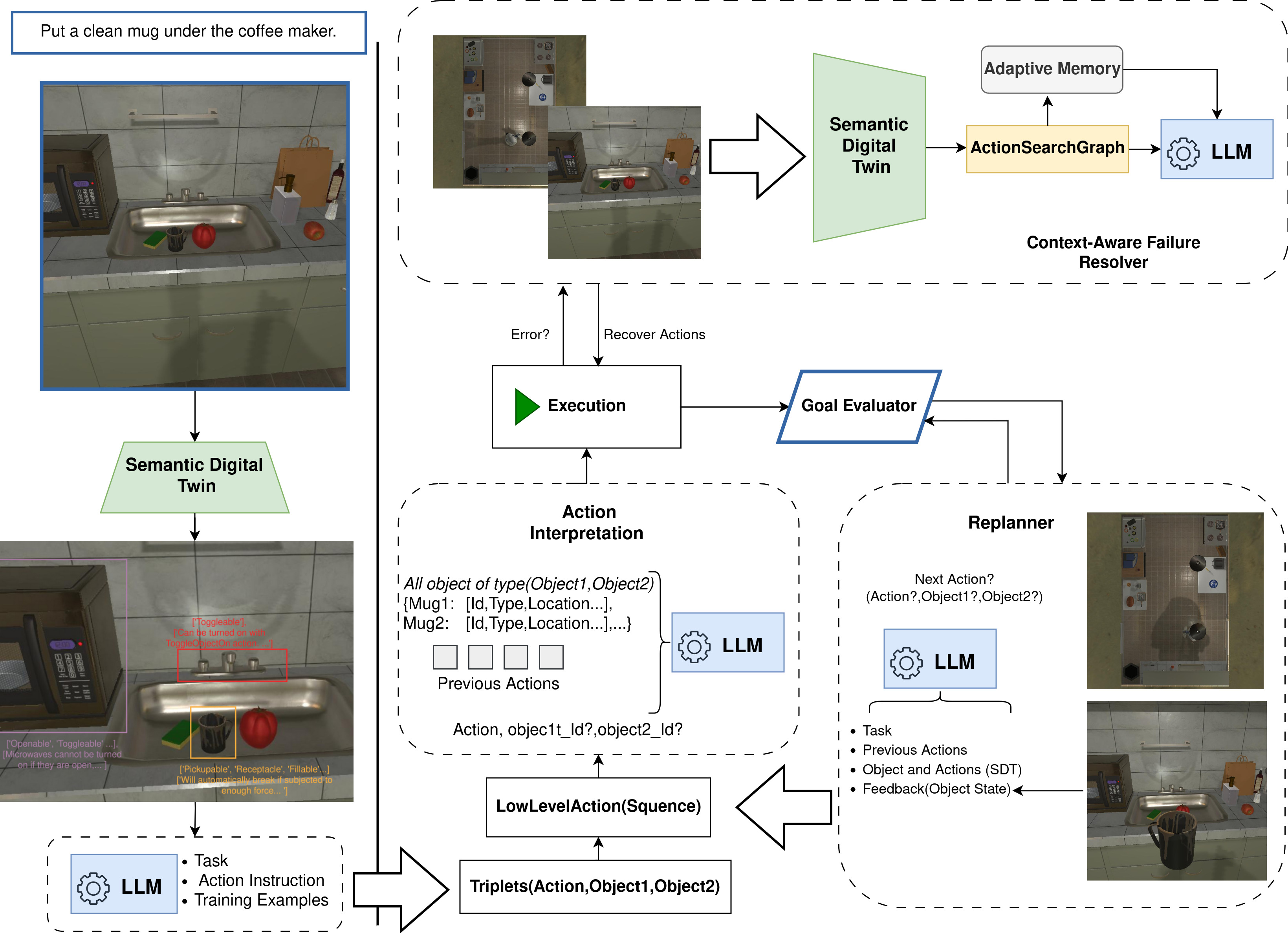}
      \caption{LLM-based task planning with a Semantic Digital Twin. The SDT provides real-time context, enabling the LLM to generate and adapt action for goal-directed planning and execution.}
      \label{methodology}
\end{figure*}

\IEEEpeerreviewmaketitle

\section{Introduction}
Exploring dynamic environments has revolutionized how robots interpret and execute complex tasks by leveraging advanced natural language processing capabilities \cite{huang2023instruct2actmappingmultimodalityinstructions}. This allows robots to decompose high-level instructions into actionable steps and adapt dynamically \cite{wang2024robogenunleashinginfinitedata}, as seen in frameworks like RePLan \cite{skreta2024replanroboticreplanningperception}\cite{xie2024humanlikereasoningframeworkmultiphases}, which enables online re-planning in response to unexpected obstacles. However, challenges such as safety concerns and model hallucinations (where models generate plausible but incorrect outputs) persist. To address these, some work incorporates safety checks into the planning process\cite{bhat2024groundingllmsrobottask}\cite{khan2025safetyawaretaskplanning}, while others use scene graphs to ground LLMs in the physical environment\cite{liu2025deltadecomposedefficientlongterm}, enhancing plan feasibility by connecting language models to real-world contexts through sensor data or computer vision.

Traditional Digital Twins \cite{melnik2025digital} primarily focus on geometric and physical representations of the environment, such as objects, meshes, and features, whereas Semantic Digital Twins (SDTs) incorporate real-time data \cite{yang2023lidarllmexploringpotentiallarge}. It mirrors a building's systems and components of DT by adding 
layer of rich contextual and semantic information about dynamic surroundings by integrating the objects' text descriptions, rules, and interaction properties\cite{gu2023conceptgraphsopenvocabulary3dscene}\cite{chen2025g3flowgenerative3dsemantic}.Various approaches have been used to SDTs with LLM agent in affordance-based scene representations to enable facilitate large-scale task planning in complex environments\cite{zeng2023largelanguagemodelsrobotics}\cite{article}.

The fusion of LLM and SDTs offers a promising path to creating robotic systems capable of operating in dynamic environments. The relevance of this collaboration is to support a robot's ability by effectively adjusting robot actions that rely on real-time environmental feedback\cite{nasiriany2024pivotiterativevisualprompting}\cite{xie2024humanlikereasoningframeworkmultiphases}. 
However, the main challenge is to align the LLM response with the semantic structures represented in SDTs, ensuring seamless communication between the two systems. Our framework addresses these challenges essential for bridging high-level reasoning with real-world execution in dynamic environments. This paper discusses the necessary preliminaries, including related work on integrating LLMs and SDTs, workflow construction, experimental setup, and evaluation results. Finally, it concludes with potential directions for future research.

\section{Related Work}
\subsection{Leveraging Semantic Digital Twins for Task Planning}
Recently, there have been advancements in how robots plan tasks, especially when they don't have all the information \cite{melnik2023uniteam, yenamandra2024towards, ps2024splatr}. AutoGPT+P \cite{Birr_2024} is an example of robots using object affordances and large language models to make plans even when certain objects are missing from the environment. However, this method can be slow because it relies on external LLM models for affordance mapping. In contrast, our approach employs Semantic Digital Twins (SDTs) that encapsulate object-action rules. This allows agents to plan tasks in real time without the need for training or external object reasoner, enhancing responsiveness and reducing reliance on potentially unreliable external data sources.

\subsection{Integrating LLMs and SDTs in Complex Environments}
Approaches like SayPlan utilize 3D scene graphs to ground LLM-generated plans in large-scale, multi-room environments\cite{rana2023sayplangroundinglargelanguage, rana2023contrastive, arjun2024cognitive}. Similarly, the SMART-LLM framework demonstrates using LLMs for multi-agent task planning, translating high-level instructions into coordinated actions\cite{kannan2024smartllmsmartmultiagentrobot}. These methods, while scalable, often depend on extensive pre-training and may struggle with real-time adaptability\cite{abe2025llmmediateddynamicplangeneration}. Our method diverges by using SDTs that inherently understand object-action relationships \cite{mikami2024natural}. These enable agents to plan and adapt in complex environments, facilitating immediate responsiveness to dynamic changes and enhancing operational efficiency.

\subsection{ Failure Detection and Recovery}
Traditional failure recovery systems in robotics often combine LLMs with SDTs to detect and adapt to errors\cite{khan2025safetyawaretaskplanning}. For example, REFLECT \cite{liu2023reflectsummarizingrobotexperiences} uses vision-language models (VLMs) \cite{driess2023palmeembodiedmultimodallanguage} conditioned on failure-related queries to generate summaries and re-plan tasks. Recent work such as LoTa-Bench has brought significant attention to evaluating language-based task planners in failure-prone environments \cite{choi2024lotabenchbenchmarkinglanguageorientedtask}. LoTa-Bench provides a benchmark to assess how well LLM-based agents detect, describe, and recover from execution failures. It emphasizes querying large models during execution to reason about error types and generate recovery strategies. While LoTa-Bench has made valuable contributions in standardizing failure scenarios and offering metrics for recovery evaluation, its reliance on external querying and LLM-based introspection introduces latency and potential inconsistency during run-time execution.

\section{Methodology}
Enabling LLMs to operate in dynamic surroundings, reacting to changing robotic states and environments without excessive computational latency, our framework leverages a Semantic Digital Twin (SDT) to prompt a Large Language Model (LLM), enabling it to decompose the natural language description of a target task into a sequence of structured action triplets. These triplets act as step-by-step execution guidelines for the robot. Next, the agent grounds each action triplet by selecting the most contextually appropriate parameters to ensure smooth execution of the corresponding trajectory. The agent also handles potential failure cases at this step by leveraging an object-action knowledge semantic map to predict suitable recovery actions.

\subsection{Semantic Digital Twin}
Our framework consists of two key Semantic Digital Twin components: ``Rules and Interaction Properties" and ``Textual Descriptions." These components represent objects' responses to specific actions based on the surrounding context and conditions. Specifically, we utilize a structured set of rules describing object affordances and the consequences of particular interactions (Figure~\ref{fig:sdt_object_rules},Figure~\ref{fig:sdt_object_actions}). For example: Bottle, ``Will break if subjected to sufficient force. Will fill up with water if placed under a running water source." We also define a structured set of permissible actions. For example, the object ``Bottle" is annotated with the action properties [``Pickupable", ``Fillable", ``Breakable"]. These affordances indicate that the bottle can be picked up, filled (e.g., with water), or broken under certain conditions. Performing such a link explicitly to objects with corresponding action capabilities enables the agent to create a concept map of a given task. By grounding the task description with a concept map of SDT using LLMs, we predicted the set of action triplets and later handled the failure cases followed by the replanner. By grounding the task description using a conceptual map derived from the Semantic Digital Twin (SDT) and leveraging the reasoning capabilities of Large Language Models (LLMs), we first predicted a structured set of action triplets for task execution and later handled the failure cases.

\begin{figure*}[thpb]
      \centering
      \includegraphics[width=0.8\textwidth]{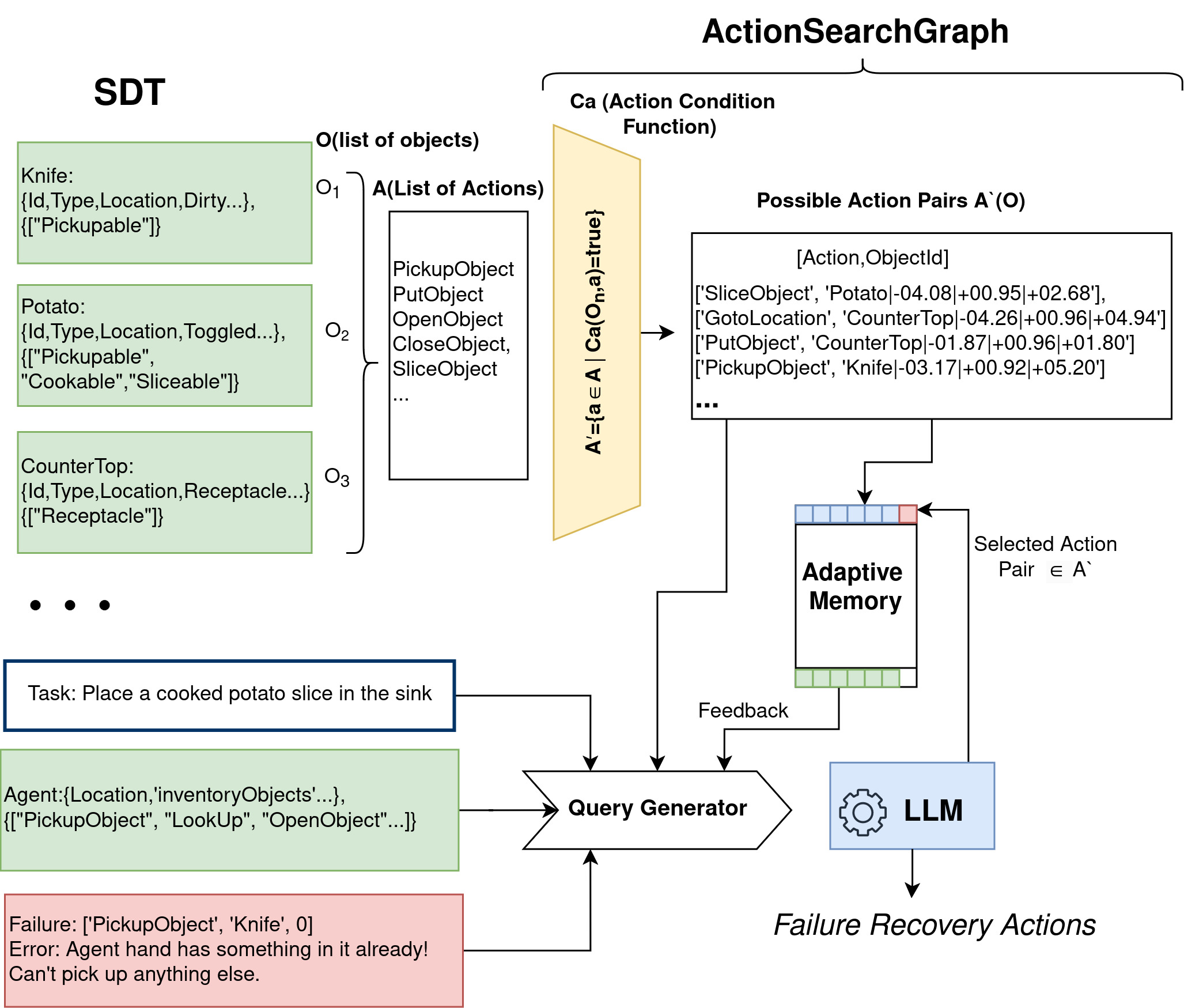}
      \caption{Overview of the Context-Aware Failure Resolver system: SDT Object descriptions are processed with a condition function to filter possible actions from the action set (A). Filtered action pairs are passed to the Adaptive Memory and Query Generator.}
      \label{failure_resolver}
\end{figure*}

\begin{figure}[!b]
      \centering
      \includegraphics[width=\linewidth]{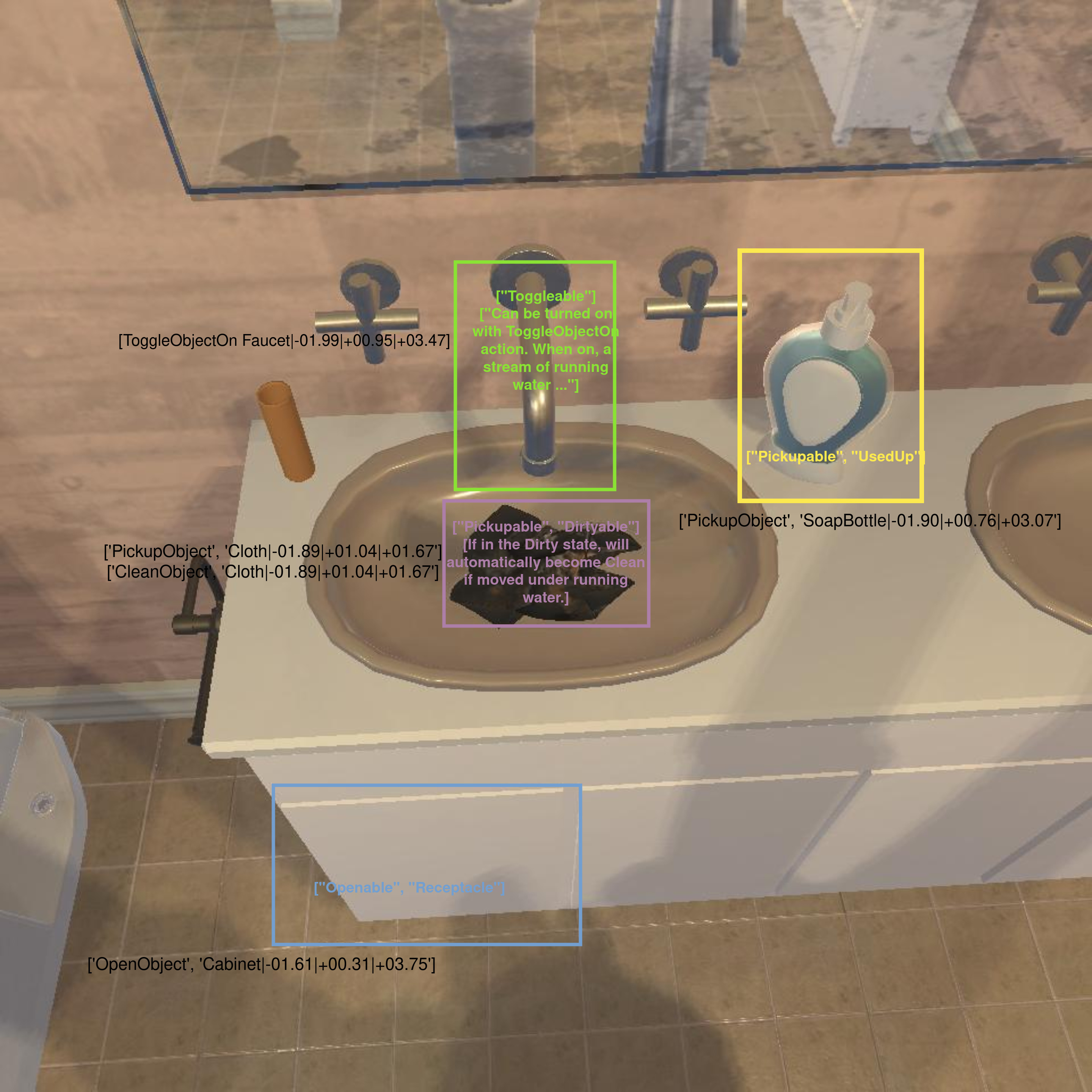}
      \caption{Semantic Digital Twin based on ``Rules and Interaction Properties" and ``Textual
Descriptions"}
      \label{SDT}
\end{figure}

\subsection{LLM-Powered Adaptive Planner}
To provide Large Language Models (LLMs) with structured input for triplet prediction, we design an LLM-based agent that operates in a multi-step process. First, the agent scans the environment and filters out only the objects that are relevant to the task at hand. Once these task-relevant objects are isolated, the agent queries a SDT, which supplies information about the possible actions that can be performed on or with each object, along with the actionable properties associated with them (e.g., whether an object is fillable, openable, or can be picked up). With this contextual understanding, comprising the filtered set of objects, their actionable properties, and the predefined rules or instructions that govern permissible actions along with previous task examples, the LLM agent is equipped to generate a structured output. Specifically, it predicts a sequence of action triplets, where each triplet \textit{Triplets(HighLevelAction,Object1,Object2)} typically represents an action, the object it applies to, and any associated parameters. The training examples are task-context examples used to construct the prompt, providing the LLM agent with guidance on the expected type of response. Alongside this sequence, the agent also predicts the final goal condition these actions aim to achieve.

After establishing general guidelines for the task in the form of triplets, the next crucial step involves resolving the specific parameters of the predicted action triplets. In this context, resolving parameters means determining the most appropriate objects to associate with each high-level action in the sequence. To accomplish this, the LLM agent performs step-by-step reasoning during action execution in the Action Interpretation Engine. At each step, the agent takes into account three key inputs: (1) the current state SDT of the environment, (2) the task description, and (3) the history of previously executed actions. Using this information, the agent constructs a context-aware query to determine which object in the environment is the most suitable for the following high-level action in the triplet sequence.

\subsection{Context-Aware Failure Handling and Replanning}
If an action fails during execution, the Failure Resolver intervenes and attempts to resolve the issue by predicting alternative actions. It explores the environment and uses Action Search Graph to generate a map of available action pairs as shown in Figure~\ref{failure_resolver} and Figure~\ref{SDT}. The Failure Resolver Engine also incorporates adaptive memory, which tracks previously attempted solutions to avoid repeating the same ones for a specific failure point.

The Context-Aware Failure Resolver is designed to analyze errors and suggest appropriate recovery actions to address the failure. Action Search Graph takes a set of object \textbf{O}, a list of actions \textbf{A}, and filters the actions based on a condition function \textbf{Ca}.

\begin{equation}
A' = \{ a \in A \mid C_a(O, a) = \text{\textit{T}} \}
\end{equation}
Here \textbf{O} be a list of object descriptions generated from SDT, \textbf{A}=[a1,a2,...,an] be a finite set of possible actions. \textbf{Ca(O,a)} is a boolean-valued condition function, which returns true if action a is valid for object \textbf{O} and false otherwise. The filtered action set \textbf{A} consists of all actions a from the original set \textbf{A} for which the condition function \textbf{Ca(O,a)} evaluates to true.
After generating all possible action pairs, they are sent to both the adaptive memory and the query generator. Adaptive memory is responsible for tracking the action pairs that have been attempted and providing feedback on them. The Query Generator then formulates a query (as illustrated in the Figure~\ref{fig:failure_prompt}) to the LLM agent to identify a potential solution. If the suggested action pair fails to resolve the error, a new query is generated, incorporating feedback from the previous attempt.

Once all predicted triplets have been executed successfully, the agent evaluates the task by comparing the current state with the goal state. If the goal has not yet been achieved, it gathers information on the actions performed so far, the current state of the objects, the task description, and feedback on the unmet goal conditions. This information is then used to prompt the LLM agent to generate additional action steps (in the form of triplets) needed to fulfill the remaining goal conditions.

\section{Experiment}
We systematically evaluate our approach using examples from the ALFRED dataset\cite{shridhar2020alfredbenchmarkinterpretinggrounded} build on ai2thor, a benchmark designed for embodied AI agents that plan and execute low-level actions to accomplish household tasks. These tasks include actions such as cleaning a mug, cutting vegetables, and cooling an apple in the refrigerator. For our evaluation, we selected tasks from three categories: Clean\&Place, Heat \& Place, and Cool \& Place,as they present greater challenges in terms of reasoning and interaction complexity. The examples are drawn from the valid-seen split, which includes various interactive actions such as SliceObject, OpenObject, and ToggleOnObject. The SDT is generated from the behavior of each type of object that appears in ai2thor. \url{https://ai2thor.allenai.org/} We use an SDT-assisted LLM to break down the language task description into triplets. Failure conditions are generated by altering the environment's ground truth from the dataset, such as making an object dirty or placing it inside a closed container.

\begin{table*}[t]
\centering
\caption{Summary of Task Execution Performance on for Various Household Tasks}
\begin{tabular}{|r|l|c|c|c|c|}
\hline
 \textbf{Task IDs} & \textbf{Task Description} & \textbf{No. Failure} & \textbf{Iteration Per Failure} & \textbf{Replanner Iteration} & \textbf{Success} \\
\hline
\rowcolor{gray!20}
1& Place a cooked potato slice in the sink     & 2 & 2 & 0 & Yes \\
2&Put a cooked piece of potato in the sink. & 0 & 0  & 2 & Yes \\

\rowcolor{gray!20}
3&Place a rinsed knife inside a drawer.       & 1 & 1    & 0 & Yes \\
4&Slice an apple, cook it and set it on the counter   & 1 & 1 & 0 & Yes \\

\rowcolor{gray!20}
5&Place a clean knife in the drawer & 0 & 0 & 0 & Yes \\
6&Put a warm apple slice on the counter.    & 2 & 2 & 0 & Yes \\

\rowcolor{gray!20}
7&To cook a sliced tomato to throw it in the trash.  & 1 & 1  & 0 & Yes \\
8&Put a chilled plate on the counter left of the sink.   & 2 & 2 & 0 & Yes \\

\rowcolor{gray!20}
9&Set a chilled bottle of wine on the table.  & 1 & 4  & 0 & Yes \\
10&Put a clean mug under the coffee maker.   & 0 & 0 & 0 & Yes \\

\rowcolor{gray!20}
11&Put a clean cloth on the back of the toilet. & 0 & 0  & 0 & Yes \\
12&Put a wet sponge on the counter.   & 1 & 2 & 0 & Yes \\

\rowcolor{gray!20}
13&Cool a slice of bread and put it in the microwave. & 0 & 0  & 0 & Yes \\
14&Cut an apple, cool a piece and bring it to the table  & 1 & 1 & 2 & Yes \\
\hline
\end{tabular}
\label{tab:model-performance}
\end{table*}

\section{Result and Discussion}
We selected a subset of 14 tasks for evaluation that are enactable by an agent. All tasks were successfully completed, demonstrating the robustness of the framework. Success was defined as achieving the specified goal condition. Although all tasks ultimately succeeded, several experienced failure cases during initial attempts. The system used an iterative correction mechanism for each failure case to refine the plan. For example, the task: Place a cooked potato slice in the sink; experienced two failure cases, each of them resolved in 1 iteration. We recorded the number of failure cases per task and the number of iterations required to resolve each failure across all tasks as shown in Table \ref{tab:model-performance}. This allowed us to assess the planner's adaptability and the LLM's ability to generate corrective actions. Replanner iteration specifies how often the agent invoked the replanner after completing the sequence of predicted triplets in cases where the goal condition was still unmet. This does not refer to replanning after individual triplet failures but after executing all planned steps and failing to reach the desired end state. A zero value means that the predicted triplets were sufficient to achieve the goal.The following Figure \ref{graph}compares subgoal success rates across three setups: using triplets generated by the planner alone, using triplets with the failure resolver, and using triplets with the failure resolver followed by the replanner.

Based on the results, we observe that most failure cases stem from incorrect triplet predictions, while others are due to inappropriate object selection. For instance in task ``Place a rinsed knife inside a drawer", the agent attempts to place a object on target but fails because the chosen target lacks sufficient space. In such scenarios, the agent needs to reconsider and revise its object selection. For tasks where the goal condition was not satisfied after initial execution, the Replanner Module was triggered. This component analyzed the current state and generated additional corrective actions using the LLM. We recorded the number of replanning iterations required per task. Notably, some tasks (such as the heating object examples) did not require any replanning, indicating that the initial plan and failure resolver were sufficient. However, if a goal condition is not met—for example, the agent performs the "heat the potato" action but misses the "slice" action, even though the goal requires the object to be both hot and sliced—then the replanner module steps in. It suggests the agent pick up a knife and perform the slicing action, thereby fulfilling the goal condition.

\begin{figure}[!b]
      \centering
      \includegraphics[width=\linewidth]{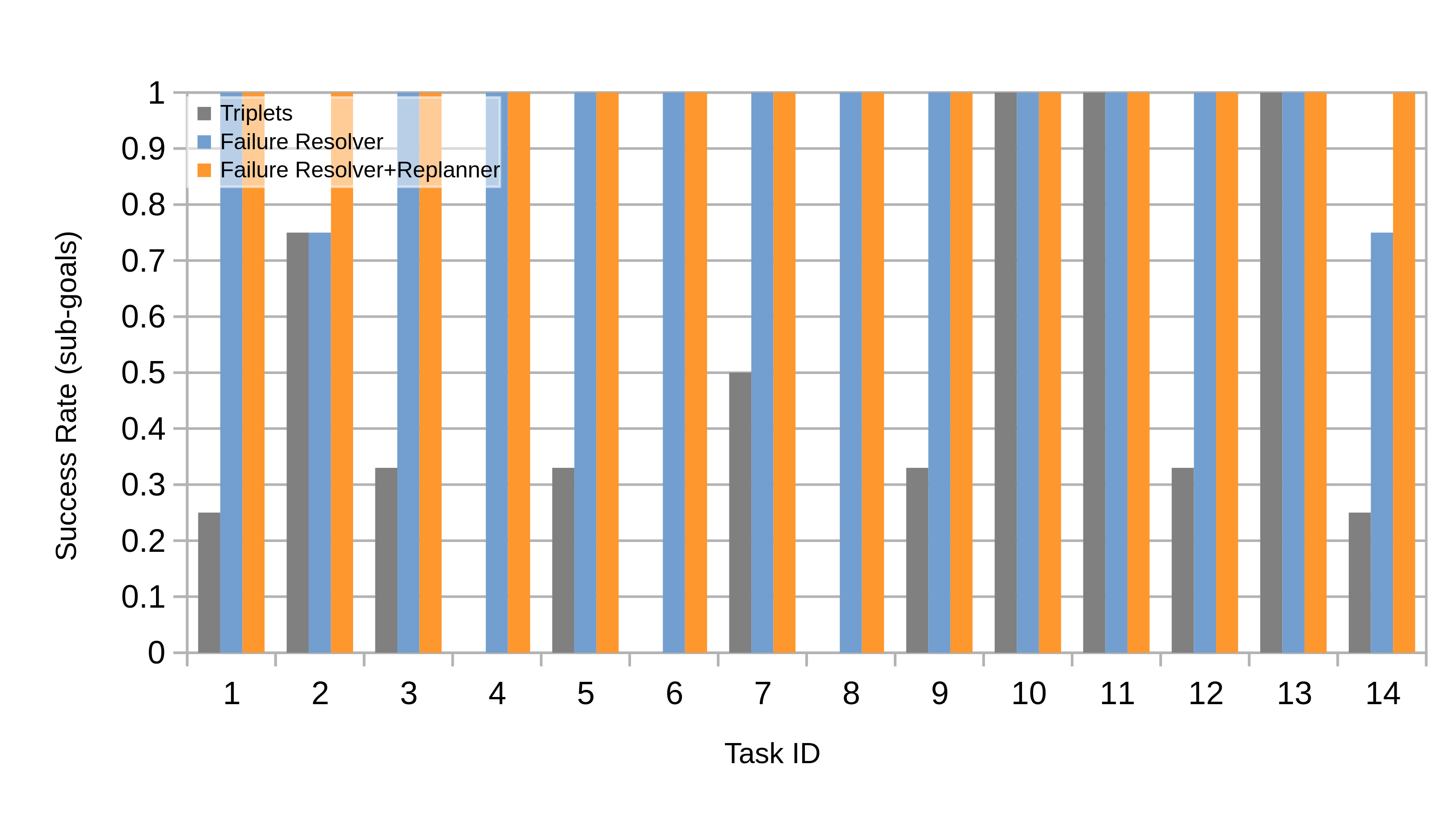}
      \caption{Comparison of task (in Table\ref{tab:model-performance}) success with and without Failure Resolver and Re-planner}
      \label{graph}
\end{figure}
The purpose of our structured "Rules and Interaction Properties" goes beyond encoding general object behavior. These rules are explicitly designed to reflect the capabilities and constraints of specific robot agents and simulation environments. For example. While an LLM may know that a bottle can be picked up, filled, or broken, the actual feasibility of these interactions depends on the agent's embodiment and the simulation's physics fidelity. A bottle labeled as "Pickupable" in our framework means that it is semantically and mechanically feasible for a particular robot in a given simulator to perform the action — accounting for reachability, gripper type, force limits, etc. Our framework captures contextual dependencies and interaction consequences that may not be static or universally known. For example, whether an object breaks may depend on the surface it's dropped on, or whether it fills depends on proximity to a simulated water source — details that require structured specification rather than reliance on general commonsense priors. Thus, the rules provide grounded, operationalizable knowledge that is critical for agents operating in simulated or real environments where affordances are not just about what "should" happen but what can happen under defined system constraints.

In earlier research, failures were typically addressed by summarizing the actions taken and then replanning the entire task from the beginning\cite{liu2023reflectsummarizingrobotexperiences}. In contrast, LOTA-Benchmark \cite{choi2024lotabenchbenchmarkinglanguageorientedtask} examined six failure cases and aimed to resolve them through demonstration-based feedback and example-driven replanning. None of these approaches consider SDT based on object-actionable properties to predict context-aware actions. Our proposed method addresses failure cases such as the absence of visual grounding (e.g., attempting to interact with invisible or non-present objects) and object selection errors (e.g., choosing the wrong object). Moreover, the Action interpretation Engine helps execute the action triplets by determining whether interacting with an object logically follows previous actions. Its purpose is to ensure that each triplet is valid on its own and contributes meaningfully to accomplishing the overall task.This is achieved through the use of a Semantic Digital Twin-based ActionSearch Graph, which enhances the agent's understanding of object properties and contextual relevance.

\section{Conclusion} 

This work introduces an adaptive framework that synergistically combines Semantic Digital Twins (SDTs) with Large Language Models (LLMs) to enhance robotic task planning and execution in dynamic, real-world environments. By representing tasks as structured action triplets grounded in semantic and contextual information, our system enables execution, real-time error recovery, and iterative replanning. Experimental evaluations using the ALFRED dataset validate the effectiveness of our approach in handling complex household tasks, even in scenarios involving uncertainty or unexpected failures. Unlike prior methods, our framework leverages the rich semantics of SDTs to inform context-aware decision-making, eliminating reliance on external affordance mapping or retraining. This integration bridges the gap between high-level language reasoning and low-level robotic control, setting a new direction for autonomous agents capable of reliable, interpretable, and resilient behavior. Future work will explore expanding the semantic representation of environments, and incorporating multimodal sensory feedback for more nuanced task understanding. 

\section*{Acknowledgments}
This research was partially funded by the German Research Foundation DFG, as part of Collaborative Research Center (Sonderforschungsbereich) “EASE - Everyday Activity Science and Engineering”, University of Bremen.

\bibliographystyle{plainnat}
\bibliography{references}

\newpage
\clearpage
\onecolumn
\section*{Appendix}
\addcontentsline{toc}{section}{Appendix} %
\begin{center}
    \fbox{%
        \includegraphics[width=0.95\textwidth]{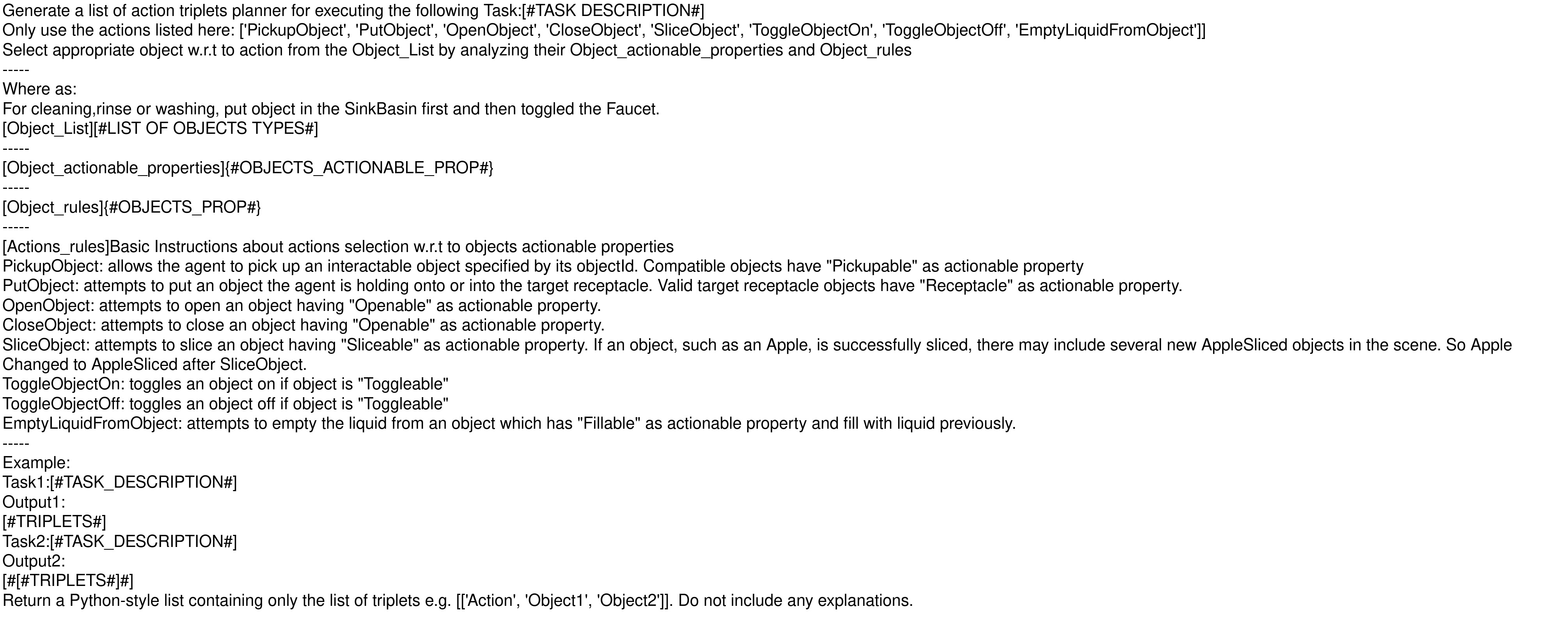}
    }
    \captionof{figure}{Action-Triplets Prompting}
    \label{fig:triplet_prompt}
\end{center}

\begin{center}
    \fbox{%
        \includegraphics[width=0.95\textwidth]{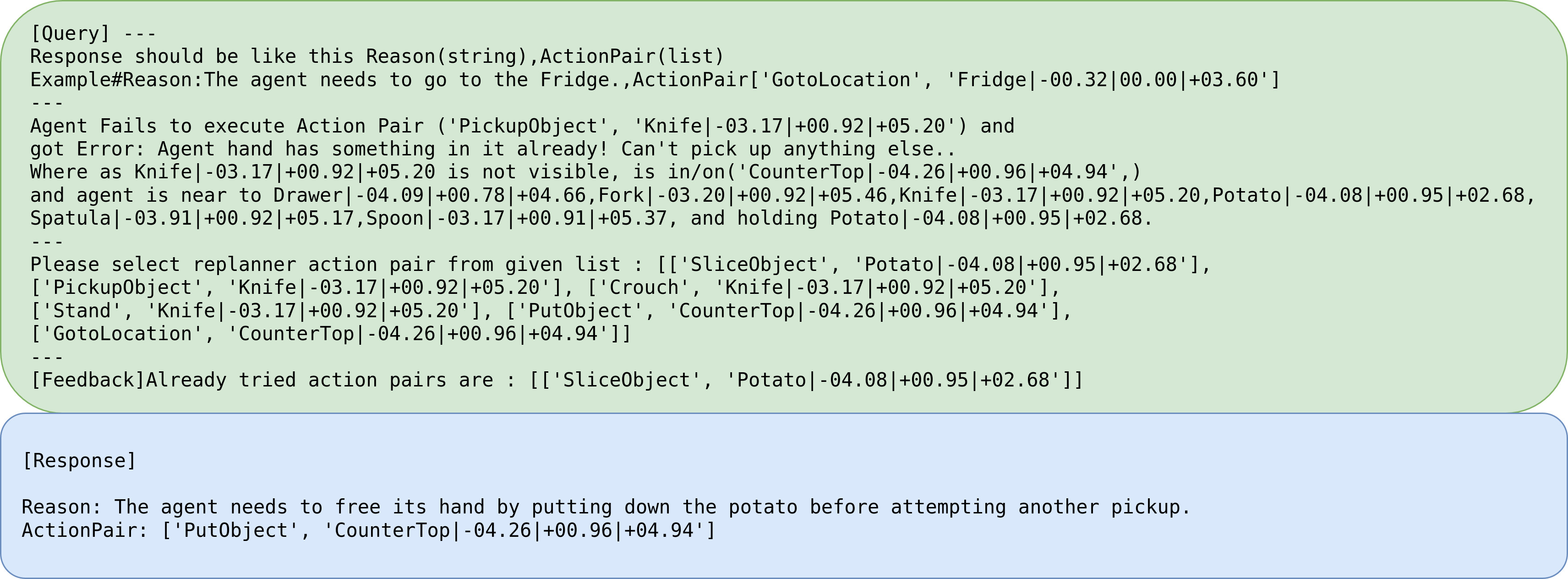}
    }
    \captionof{figure}{Example of Failure Query}
    \label{fig:failure_prompt}
\end{center}
\newpage
\clearpage

\begin{center}
    \fbox{%
        \includegraphics[width=0.95\textwidth]{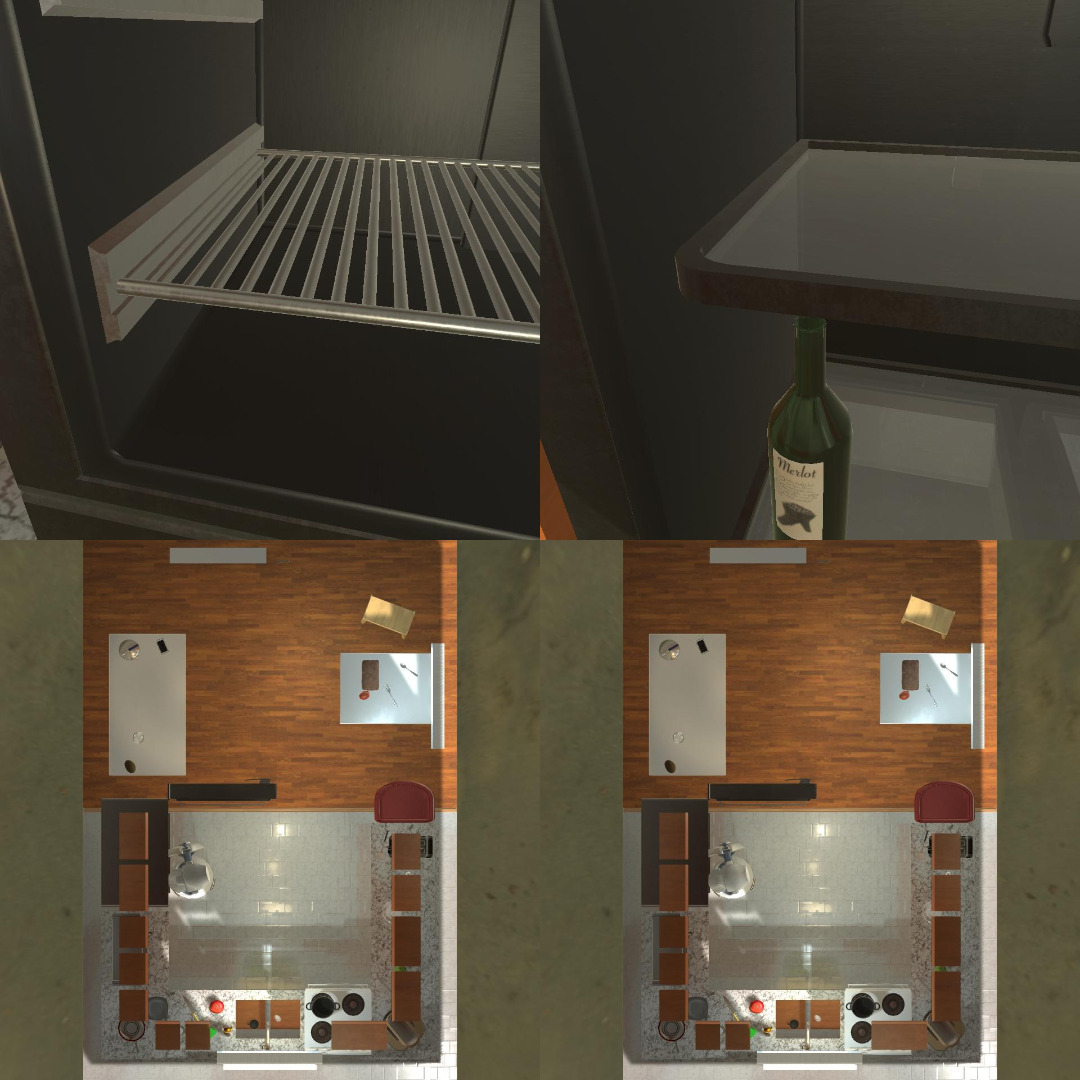}
    }
    \captionof{figure}{Robot fails to locate bottle in a Fridge(top-left) then it locate the bottle (top-right) after Failure resolver}
    \label{fig:eg_bottle}
\end{center}

Task: Set a chilled bottle of wine on the table. Action-Triplets:[['PickupObject', 'WineBottle', 0], ['OpenObject', 'Fridge', 0], ['PutObject', 'WineBottle', 'Fridge'], ['CloseObject', 'Fridge', 0], ['OpenObject', 'Fridge', 0], ['PickupObject', 'WineBottle', 0], ['CloseObject', 'Fridge', 0], ['PutObject', 'WineBottle', 'DiningTable']]. Agent got an Error: ``Target object not found within the specified visibility..." during ['PickupObject', 'WineBottle', 0] Failure Resolver suggested solution actions are: [(Crouch,Fridge$\mid$ -01.30$\mid$+00.01$\mid$+00.99),(PickupObject,WineBottle$\mid$-01.38$\mid$+00.76$\mid$+02.20)]

\newpage
\clearpage
\begin{figure}[h]
    \centering
    \fbox{%
        \includegraphics[width=0.3\columnwidth]{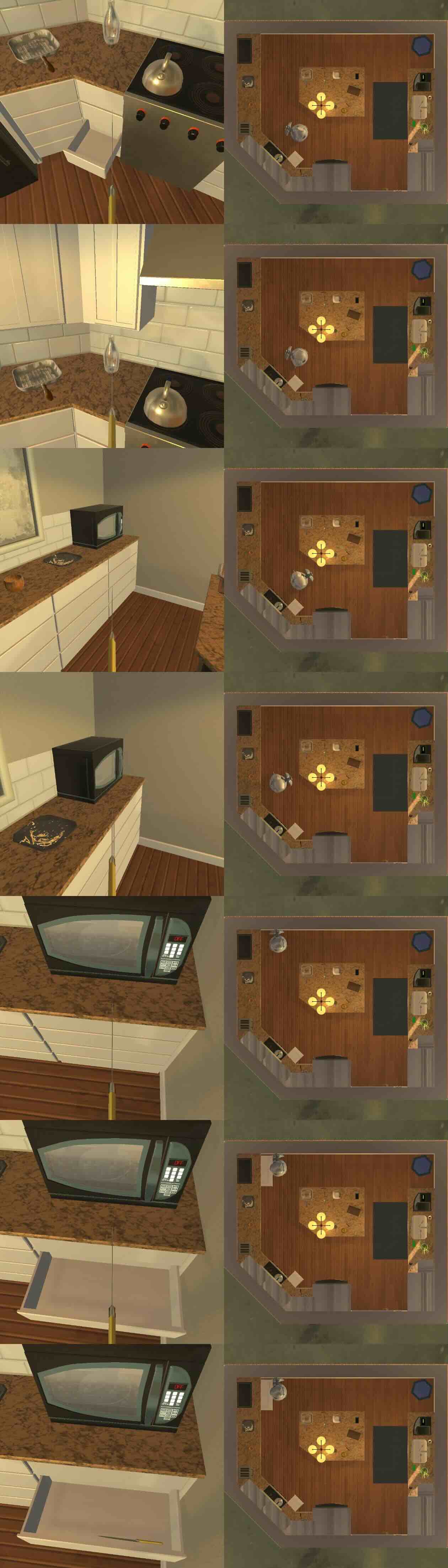}
    }
    \caption{Robot tries to put knife in a drawer.}
    \label{fig:eg_knife}
\end{figure}

TASK : Place a rinsed knife inside a drawer. Robot tries to put a Knife in a small-size drawer with the action (PutObject,Drawer$\mid$-00.05$\mid$+00.38$\mid$-01.32) but encounter a Error: ``No valid positions to place object found." Failure Resolver suggested solution actions are: [(OpenObject,Drawer$\mid$-00.86$\mid$+00.58$\mid$+01.43),PutObject,Drawer$\mid$-00.86$\mid$+00.58$\mid$+01.43)]

\newpage
\clearpage
\begin{center}
    \fbox{%
        \includegraphics[width=0.95\textwidth]{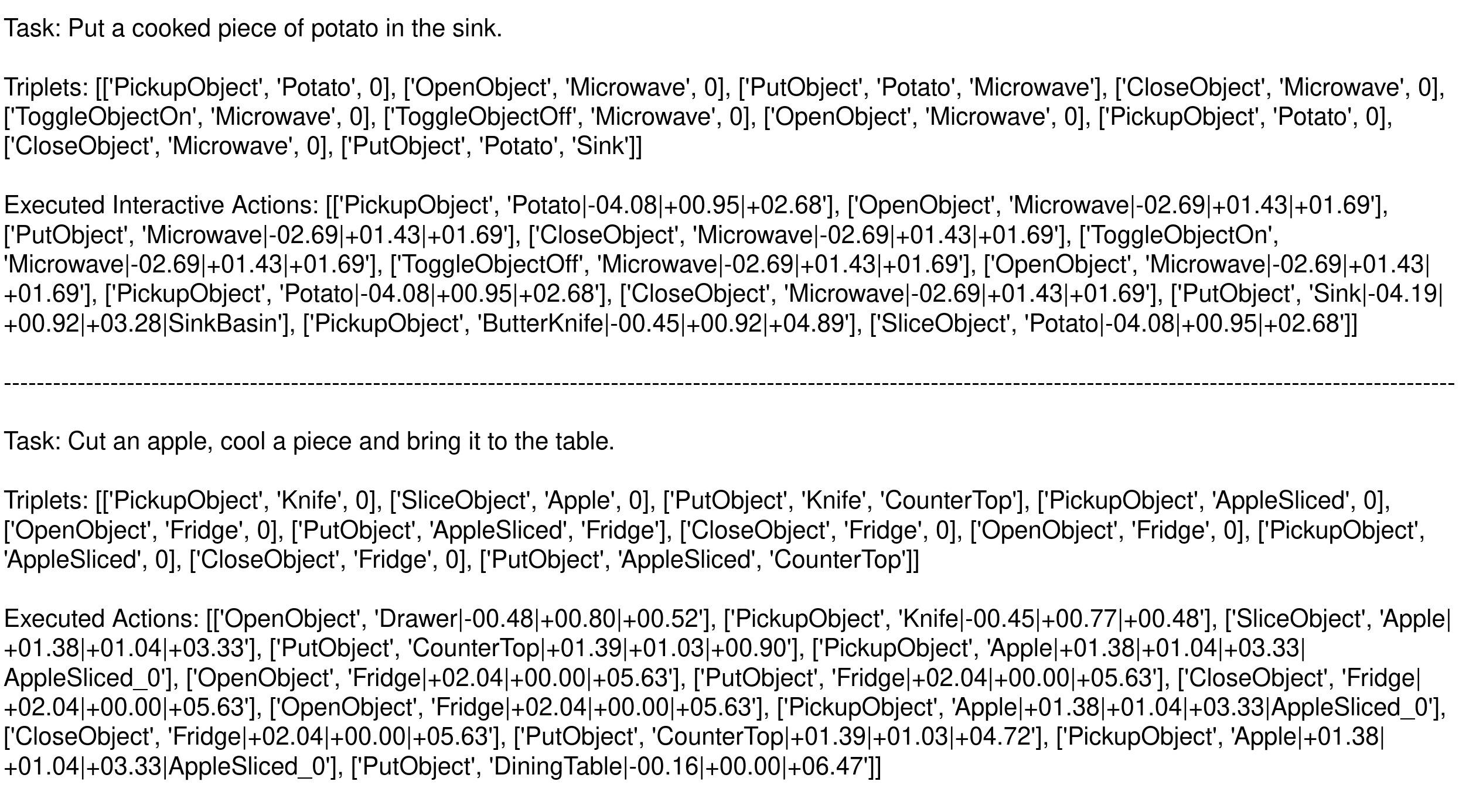}
    }
    \captionof{figure}{Tasks, along with their predicted triplets and summary of actions, leads to success after failure resolver and replanner. }
    \label{fig:eg_replanner}
\end{center}

In the case of Task: Put a cooked piece of potato in the sink, the robot heats the potato and puts it in the sink but forgets to slice it. So, the replanner module includes two more actions(['PickupObject', 'ButterKnife$\mid$-00.45$\mid$+00.92$\mid$+04.89'], ['SliceObject', 'Potato$\mid$-04.08$\mid$+00.95$\mid$+02.68']) at the end of the task to achieve the goal condition. Whereas in Task:
Cut an apple, cool a piece and bring it to the table, When robot try to execute actions-triplet ['PickupObject', 'Knife', 0] it encounter the Error: ``Target object not found within the specified visibility..". In this case, the Context-Awareness Failure Resolver predicted that the drawer would open first as the knife was in the drawer. The task’s goal is for the apple slice to be cold and on
the table, but the robot puts the apple slice on the countertop. Replanner suggested the following two actions:['PickupObject', 'Apple+01.38$\mid$+01.04$\mid$+03.33$\mid$AppleSliced-0'], ['PutObject', 'DiningTable$\mid$-00.16$\mid$+00.00$\mid$+06.47'] to achieve the goal.
\newpage
\clearpage
\begin{center}
 
        \includegraphics[width=0.95\textwidth]{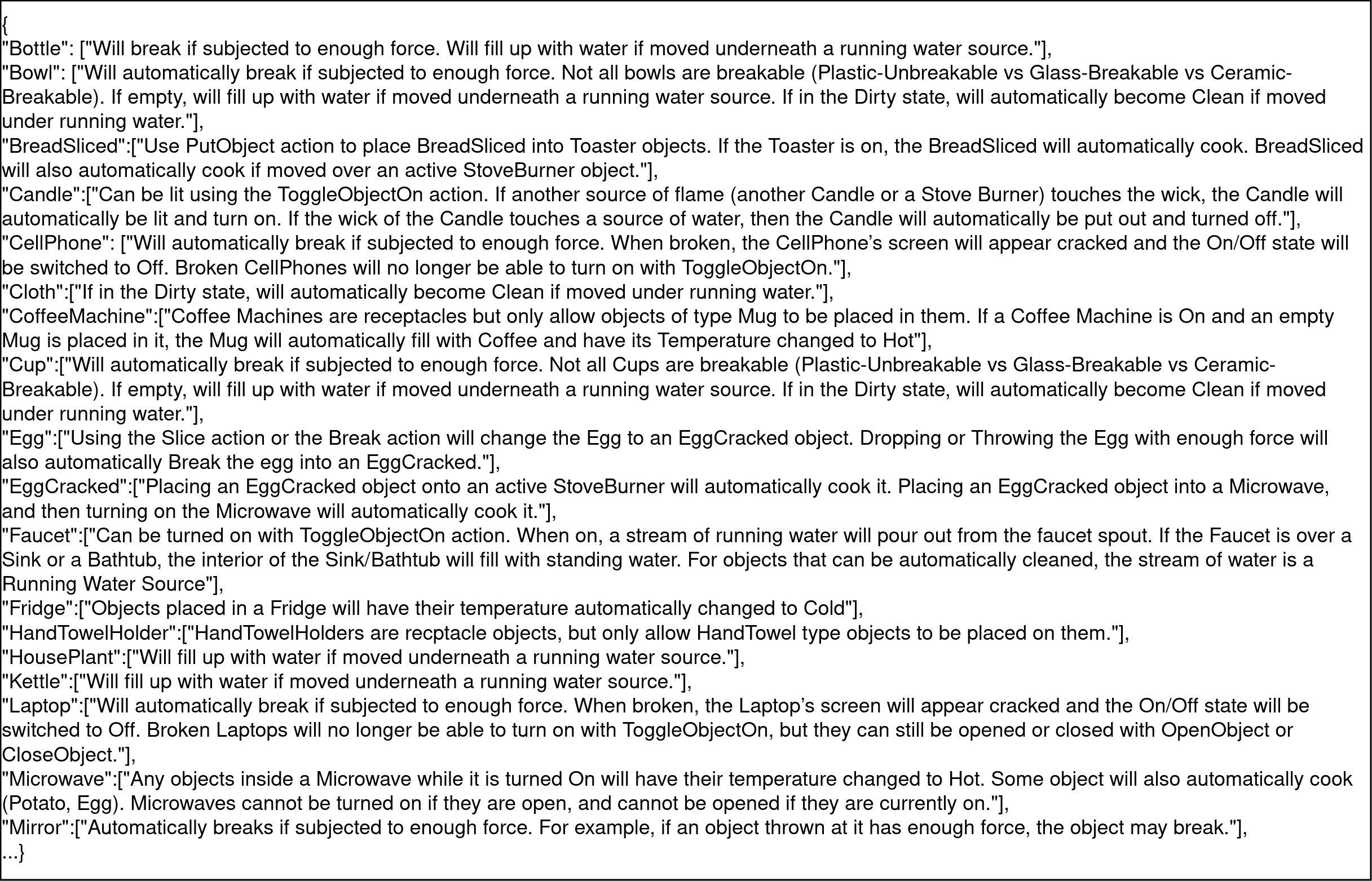}
    
    \captionof{figure}{Objects contextual interactions properties from ai2Thor \url{https://ai2thor.allenai.org/}}
    \label{fig:sdt_object_rules}
\end{center}

\begin{center}
 
        \includegraphics[width=0.6\textwidth]{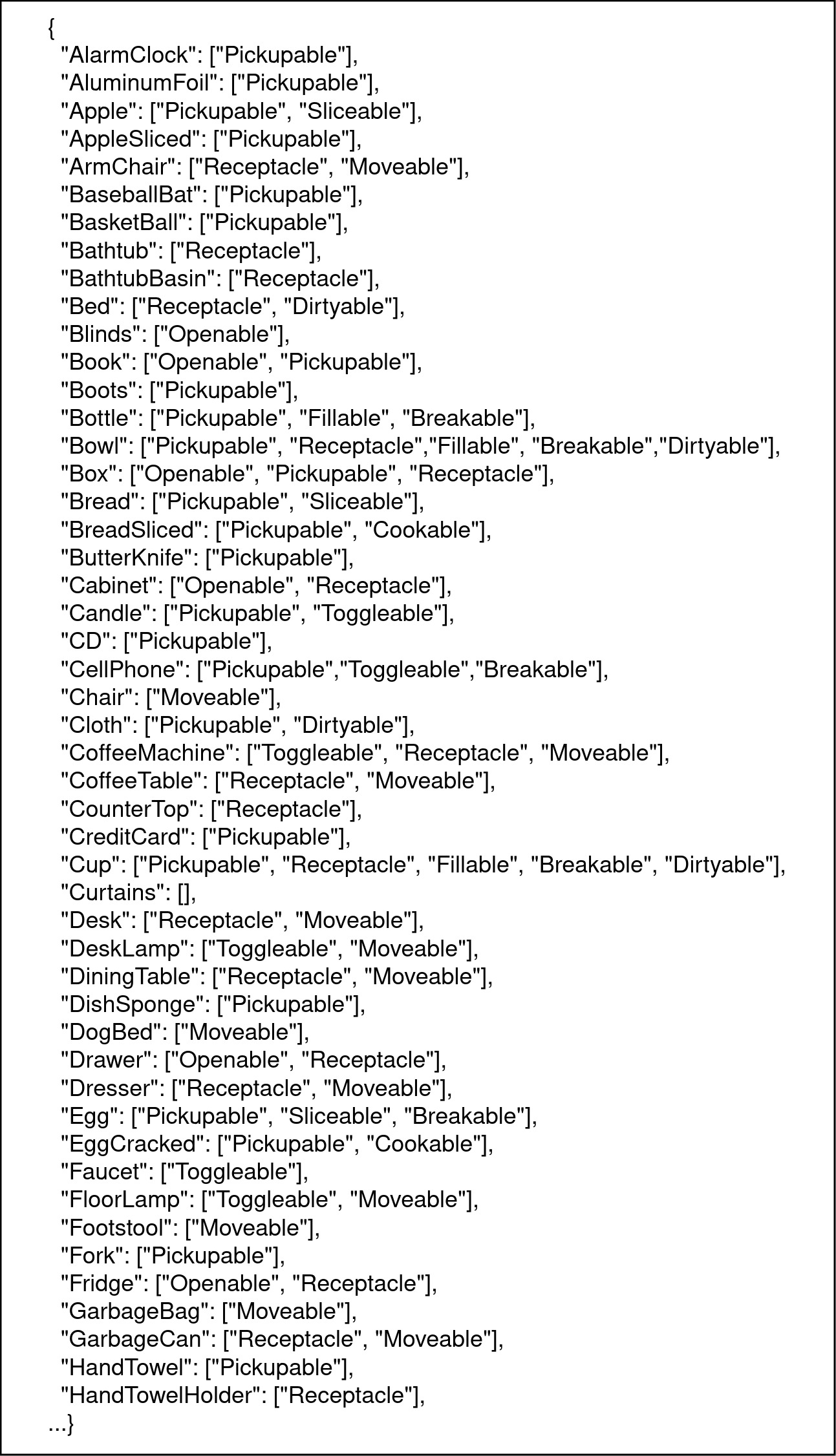}
    
    \captionof{figure}{Objects actionable properties from ai2Thor \url{https://ai2thor.allenai.org/}}
    \label{fig:sdt_object_actions}
\end{center}
\end{document}